\documentclass[runningheads]{llncs}

\usepackage[mobile]{eccv}

\usepackage{eccvabbrv}

\usepackage{graphicx}
\usepackage{booktabs}

\usepackage[accsupp]{axessibility}  %

\usepackage{hyperref}

\usepackage{orcidlink}
\usepackage{multirow}
\usepackage{makecell}
\usepackage{float}
\usepackage[numbers]{natbib}
\usepackage{enumitem}
\usepackage[table]{xcolor}
\usepackage{pifont}
\newcommand{\cmark}{\textcolor{green!60!black}{\ding{51}}}
\newcommand{\xmark}{\textcolor{red!70!black}{\ding{55}}}
\usepackage{setspace}
\usepackage{amsmath}
\usepackage{amssymb}
\newcommand{\customfootnotetext}[2]{{%
  \renewcommand{\thefootnote}{#1}%
  \footnotetext[0]{#2}}}%

\begin{document}

\title{Action Images: End-to-End Policy Learning \\via Multiview Video Generation}

\author{Haoyu Zhen\inst{1}$^*$ \and
Zixian Gao\inst{1}$^*$$^\dag$ \and
Qiao Sun\inst{1}$^\dag$ \and
Yilin Zhao\inst{2} \and \\[3pt]
Yuncong Yang\inst{1} \and
Yilun Du\inst{3} \and
Pengsheng Guo\inst{4} \and \\[3pt]
Tsun-Hsuan Wang\inst{4} \and
Yi-Ling Qiao\inst{4} \and
Chuang Gan\inst{1}
}

\authorrunning{H.~Zhen et al.}

\institute{$^1$UMass Amherst \quad $^2$NVIDIA \quad $^3$Harvard University \quad $^4$Genesis AI
}

\maketitle
\customfootnotetext{*}{Equal contribution.}
\customfootnotetext{$\dag$}{This work was done when two of the authors were remote interns at UMass.}

\begin{center}
    \vspace{-5mm}
    \textcolor{blue!70!black}{\url{https://ActionImages.github.io}}
    \vspace{-6mm}
\end{center}

\begin{figure}
    \centering
    \begin{minipage}{0.9\linewidth}
        \centering
        \includegraphics[width=\linewidth]{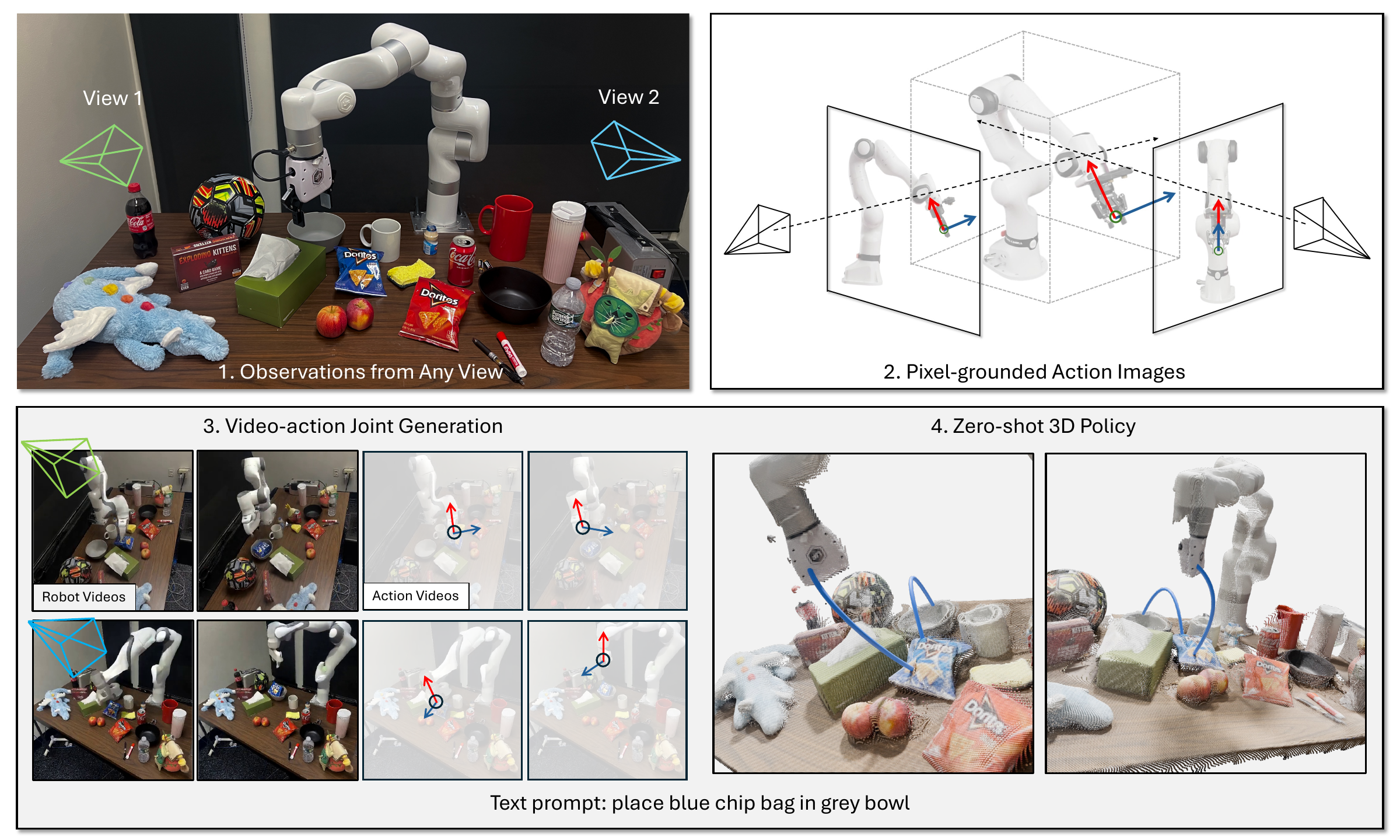}
        \caption{\textbf{Action Images} turns policy learning as multiview video generation: 7-DoF actions are translated into pixel-grounded action images that explicitly track robot-arm motion, enabling a zero-shot policy directly from a unified video backbone}
        \label{fig:placeholder}
    \end{minipage}
    \vspace{-0.6cm}
\end{figure}

\begin{abstract}
World action models (WAMs) have emerged as a promising direction for robot policy learning, as they can leverage powerful video backbones to model the future states.
However, existing approaches often rely on separate action modules, or use action representations that are not pixel-grounded, making it difficult to fully exploit the pretrained knowledge of video models and limiting transfer across viewpoints and environments. In this work, we present Action Images, a unified world action model that formulates policy learning as multiview video generation. Instead of encoding control as low-dimensional tokens, we translate 7-DoF robot actions into interpretable action images: multi-view action videos that are grounded in 2D pixels and explicitly track robot-arm motion. This pixel-grounded action representation allows the video backbone itself to act as a zero-shot policy, without a separate policy head or action module. Beyond control, the same unified model supports video-action joint generation, action-conditioned video generation, and action labeling under a shared representation. On RLBench and real-world evaluations, our model achieves the strongest zero-shot success rates and improves video–action joint generation quality over prior video-space world models, suggesting that interpretable action images are a promising route to policy learning.
\end{abstract}

\section{Introduction}

World action models~\cite{hu2024video, li2025unified, zhen2025tesseract, kim2026cosmos,ye2026world} have made rapid progress in predicting future observations, but turning this predictive ability into policy generalization remains an open challenge. In particular, strong video generation does not automatically produce a strong policy: a model may successfully synthesize plausible future frames, yet still fail to decide how to act in unseen environments. This gap between video generalization and policy generalization is a central bottleneck for world models.

A key reason is that action is still not represented in a form that world models can naturally generalize. Existing approaches typically follow one of two paths. Some~\cite{du2023learning,zhou2024robodreamer,zhen2025tesseract, ye2026world, li2025unified} attach a separate policy head or action module on top of a world model, asking an additional network to decode control from learned video features. Others~\cite{kim2026cosmos} adapt video models to action generation using representations that are not spatially grounded in image space. In both cases, the model's predictive knowledge of the world is only indirectly connected to acting. As a result, the burden of generalization is shifted to a specialized control moduel, which is often exactly where transfer breaks down.

In this work, we formulate policy learning as video generation and address policy generalization at the representation level. We propose multi-view action videos, a robotics world modeling framework that translates robot actions into interpretable action images and models them together with observations in a unified video-space representation of observation and action. Instead of treating 7-DoF control as low-dimensional signals or latent action codes, we convert each action into a pixel-grounded action representation that explicitly tracks robot-arm motion in image space across multiple views. This design makes action native to the video model itself: the same video backbone can observe, predict, condition on, and generate action, enabling a zero-shot policy. By grounding action in pixels rather than in an external interface, we obtain a more generalizable policy model that transfers more naturally across viewpoints and embodiments.

A key design choice is to represent these action images as multi-view videos. The motivation is not merely to add more visual observations, but to bridge the gap between 2D image and the 7-DoF robot action in the 3D space. A single view often provides only a ambiguous projection of motion, making it difficult for the model to infer the full action consistently from pixels alone~\cite{zhu2025aether, zhen2025tesseract}. Using multiple views makes the pixel-grounded action image more reconstructable, while also improving robustness when some motion is partially occluded.

Beyond control, the same unified video-space representation of observation and action supports multiple tasks within a single model. Because observation and action share the same generative space, the model can perform video-action joint generation, action-conditioned video generation, and action labeling under one backbone and one training objective. These capabilities emerge without a separate policy head or action module, showing that a robotics world model can be trained not only to predict the world, but also to act in it through a common visual representation.

In summary, our contributions are as follows:
\begin{itemize}
    \item We identify the gap between video generalization and policy generalization as a central limitation of current robotics world models, and argue that this gap can be addressed at the level of action representation.
    \item We propose multi-view action representation, which translate robot control into interpretable action images forming a pixel-grounded action representation, and use this representation to build a zero-shot policy without a separate policy head or action module.
    \item We show that this design yields a more generalizable policy model and provides a unified video-space representation of observation and action that supports video-action joint generation, action-conditioned video generation, and action labeling within a single robotics world model.
\end{itemize}

\section{Related Work}
\subsection{Robotics World Models.}
Originating from Reinforcement Learning~\citep{sutton1991dyna, ljung1994modeling}, world models typically take actions and the current state as input and predict future states~\citep{bruce2024genie, assran2025v}.
In recent years, learning world models for diverse robotic applications~\citep{wu2023daydreamer, zhen20243d, guo2025ctrl, bar2025navigation, li2025robotic} has garnered significant interest.
With the success of video generation models, lots of work has developed robotics world models based on video generation~\citep{du2023learning, ko2023learning, zhou2024robodreamer, videoworldsimulators2024, zhen2025tesseract, guo2025flowdreamer, sun2025learning}. These video-based approaches typically adopt a two-stage pipeline, where future observations are first predicted and actions are then generated based on these predictions.
More recently, joint video-action generation has been explored to unify modeling and control~\citep{li2025unified, kim2026cosmos}. 
In particular, DreamZero~\citep{ye2026world} demonstrates strong zero-shot generalization and cross-embodiment transfer.
However, these methods encode actions with additional action modules, leaving much of the pretrained video knowledge underused; we instead use multi-view action images so the backbone itself is a zero-shot policy.
Concurrent work~\citep{li2026multiview} also investigates video-based formulations for robot policy learning. Our approach differs in representing actions as pixel-grounded multi-view images that encode full 7-DoF control, enabling a unified video-action space and eliminating the need for separate modules.

\subsection{Generalist Robot Policy Models.}
Policy models map current states to future actions~\citep{polydoros2017survey, watkins2021explaining}. Developing generalist control policies that can succeed in diverse tasks and can be lightweightly fine-tuned to adapt to downstream tasks has long been a central goal~\citep{gupta2018robot, brohan2022rt, nakamoto2024steering, team2024octo, xing2025shortcut, zhang2025effective, li2025bfm}. 
While multiple advances in Vision-Language-Action (VLA) models~\citep{zitkovich2023rt, kim2024openvla, black2024pi_0, intelligence2025pi_}, Diffusion Policy~\citep{pearce2023imitating, chi2025diffusion}, and Reinforcement Learning~\citep{nakamoto2024steering, intelligence2025pi} have greatly promoted the generalizability of policy models, their diversity is still limited to relatively narrow task distributions and they struggle to zero-shot generalize to new environments~\citep{zheng2024towards, etukuru2025robot}. 
In parallel, strong capabilities of video generation foundation models in predicting future frames and modeling physical dynamics have inspired policy learning approaches~\citep{hu2024video, liang2025video, li2025unified}.
However, how to turn video prediction into transferable control remains nontrivial; our action-frame representation bridge this gap by making action native to the video space.

\subsection{4D Generation Models.}
``4D'' here refers to 3D plus time.
Optimization-based methods employ Score Distillation Sampling, which distills pre-trained diffusion models into specific 4D representations~\cite{singer2023text, ren2023dreamgaussian4d, yang2024deformable, bahmani20244d}. Recent work~\citep{li2025ss4d} explores native 4D generation, which is trained directly on 4D datasets.
Due to the lack of large-scale pretraining assets, a branch of research leverages the rich semantic priors in pretrained video generation models and integrates reconstruction methods to lift 2D frame sequences into 4D results~\citep{xie2024sv4d, wu2025geometry, jin2025diffuman4d, pan2024efficient4d, zhang20244diffusion}. However, these contributions mostly focus on single-avatar or simple scene generation.
Close to our method, \citep{wu2024cat4d, bai2025recammaster} leverage multiview generation to produce complex dynamic 4D scenes that can be replayed at any specified camera pose and timestamp.
However, for robotic tasks, 4D generation is typically limited to a fixed single view~\citep{zhen2025tesseract, zhu2025aether, guo2025flowdreamer}. Although \citep{liu2025geometry} has leveraged multi-view inputs and introduced a geometry-consistent supervision, they still do not generalize well beyond their training scenes.

\section{Method}

Robotics world models have recently shown strong capability in modeling dynamics, especially when built on large pretrained video backbones.
However, these advances in video prediction do not directly translate into strong policy generalization. To address this limitation, we build a unified video-space representation of observation and action, where robot control is translated into interpretable action images that form a pixel-grounded representation. We first introduce how 7-DoF robot actions are converted into multi-view action videos (Sec.~\ref{sec:action_frames}), then describe how this representation can be decoded back into continuous control with only minor information loss (Sec.~\ref{sec:action_decode}), and finally present the training of a unified world-action model that enables a zero-shot policy (Sec.~\ref{sec:train_uawm}).

\subsection{Action as Images}
\label{sec:action_frames}

\begin{figure}[tbp]
    \centering
    \includegraphics[width=0.95\linewidth]{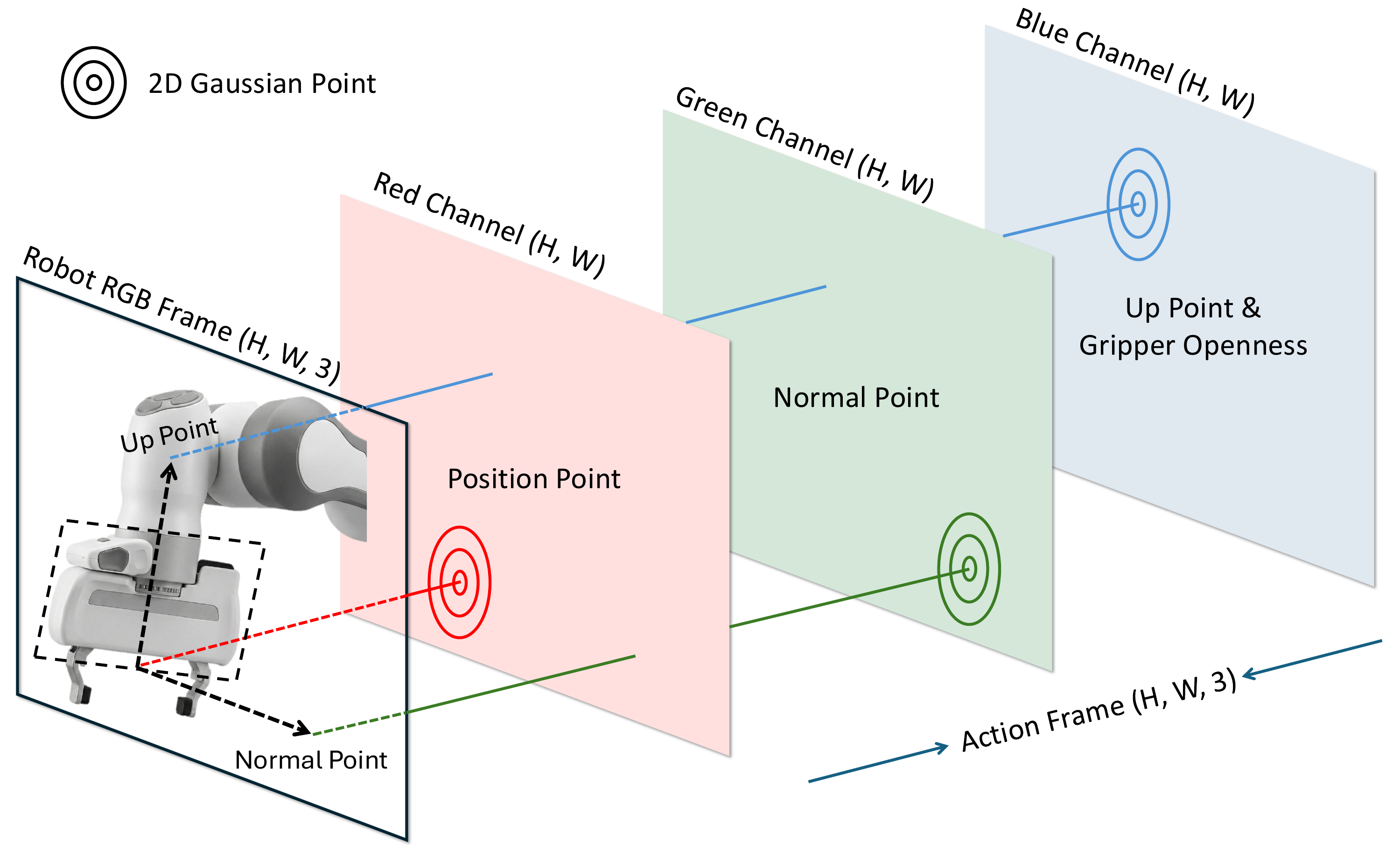}
    \caption{\textbf{Action as image.} We convert each 7-DoF robot action into three semantic 3D points (position, normal, and up), project them into image space, and render them as RGB Gaussian heatmaps. The blue channel further encodes gripper openness in the low-response background, producing a pixel-grounded action representation.}
    \label{fig:action_frame}
\end{figure}

Our central idea is to represent robot action in the same modality as robot observation. Instead of treating action as a low-dimensional control vector that must be interpreted by a separate policy head, we convert each action into interpretable action images and model it directly in video space. This yields a pixel-grounded action representation that is aligned with the robot RGB stream and can therefore be processed by the same video backbone. As illustrated in Fig.~\ref{fig:action_frame}, this design turns action modeling into a tracking-like visual prediction problem: the model does not need to infer control from abstract tokens, but instead learns to localize and reason about robot-arm motion.

\noindent\textbf{From 7-DoF action to semantic 3D points.}
At each time step $t$, the robot action is
$
\mathbf{a}_t = [\mathbf{p}_t,\ \boldsymbol{\theta}_t,\ g_t] \in \mathbb{R}^{7}
$,
where $\mathbf{p}_t \in \mathbb{R}^3$ is the end-effector position, $\boldsymbol{\theta}_t \in \mathbb{R}^3$ denotes its orientation, and $g_t \in \mathbb{R}$ is the gripper openness. We convert this 7-DoF action into three semantic 3D points: a position point, a normal point, and an up point. The position point is the end-effector position, $\mathbf{q}^{\text{pos}}_t = \mathbf{p}_t$.
The other two points are defined by rotating two canonical axes attached to the end-effector and extending them by a small length $\ell$:
\begin{equation}
\mathbf{q}^{\text{up}}_t = \mathbf{p}_t + \ell \, \mathbf{R}(\boldsymbol{\theta}_t)\mathbf{e}_x,
\qquad
\mathbf{q}^{\text{normal}}_t = \mathbf{p}_t + \ell \, \mathbf{R}(\boldsymbol{\theta}_t)(-\mathbf{e}_z),
\end{equation}
where $\mathbf{R}(\boldsymbol{\theta}_t)\in SO(3)$ is the rotation matrix derived from the action orientation. Here, the up point follows a canonical in-plane direction of the gripper, while the normal point follows the direction normal to the robot gripper plane. Together, these three points capture end-effector pose in a form that can be directly projected into image space.

\noindent\textbf{Multi-view action image rendering.}
Given a camera view $v$, we project the three semantic 3D points into image space using the camera intrinsics and extrinsics. Denoting the corresponding projection function by $\pi_t^{(v)}(\cdot)$, we obtain
\begin{equation}
\mathbf{u}^{\text{pos},(v)}_t = \pi^{(v)}_t(\mathbf{q}^{\text{pos}}_t),\quad
\mathbf{u}^{\text{normal},(v)}_t = \pi^{(v)}_t(\mathbf{q}^{\text{normal}}_t),\quad
\mathbf{u}^{\text{up},(v)}_t = \pi^{(v)}_t(\mathbf{q}^{\text{up}}_t).
\end{equation}
We then render these projected points into an action image $\mathbf{A}^{(v)}_t \in \mathbb{R}^{H\times W\times 3}$ using 2D Gaussian. The red channel encodes the position point, the green channel encodes the normal point, and the blue channel encodes the up point together with the gripper openness, as shown in Fig.~\ref{fig:action_frame}. Let
$\mathcal{G}(\mathbf{x};\mathbf{u},\sigma)$
denote a 2D Gaussian centered at pixel $\mathbf{u}$. The red and green channels are defined as
\begin{equation}
\mathbf{A}^{(v)}_t(:,:,1) = \mathcal{G}(\cdot;\mathbf{u}^{\text{pos},(v)}_t,\sigma),\quad
\mathbf{A}^{(v)}_t(:,:,2) = \mathcal{G}(\cdot;\mathbf{u}^{\text{normal},(v)}_t,\sigma).
\end{equation}
For the blue channel, we first render the up point as a Gaussian map,
\begin{equation}
\tilde{\mathbf{A}}^{(v)}_t(:,:,3) = \mathcal{G}(\cdot;\mathbf{u}^{\text{up},(v)}_t,\sigma),
\end{equation}
and then inject the binary gripper openness signal into low-response regions:
\begin{equation}
\mathbf{A}^{(v)}_t(i,j,3) =
\begin{cases}
\tilde{\mathbf{A}}^{(v)}_t(i,j,3), & \tilde{\mathbf{A}}^{(v)}_t(i,j,3) > 0.25,\\
0.25 \cdot g_t, & \text{otherwise},
\end{cases}
\end{equation}
In this way, the blue channel preserves the projected up point while also encoding gripper openness in a simple and spatially consistent form. The resulting image is an interpretable action image.
Stacking these frames over time yields an action video for each view,
\begin{equation}
\mathcal{A}^{(v)} = \{\mathbf{A}^{(v)}_1,\dots,\mathbf{A}^{(v)}_T\}
\in \mathbb{R}^{T\times H\times W\times 3}.
\end{equation}
Since these action videos have the same spatial and temporal structure as the corresponding robot RGB observations $\mathcal{O}^{(v)} \in \mathbb{R}^{T\times H\times W\times 3}$, they naturally form a unified video-space representation of observation and action.

\noindent\textbf{Benefits.}
Representing actions as interpretable action images provides two key benefits. First, it makes action prediction spatially grounded: the model learns control through visible robot-arm motion rather than through abstract action tokens. Second, it is naturally compatible with pretrained video backbones, allowing the same model to reason over observation and action without an action module. In this way, our zero-shot policy is obtained by turning the robot action into a visual prediction problem. Because the representation is pixel-grounded and multi-view, it transfers more naturally across viewpoints, motion patterns, and robot embodiments, leading to a more generalizable policy model.

\begin{figure}[t]
    \centering
    \includegraphics[width=0.92\linewidth]{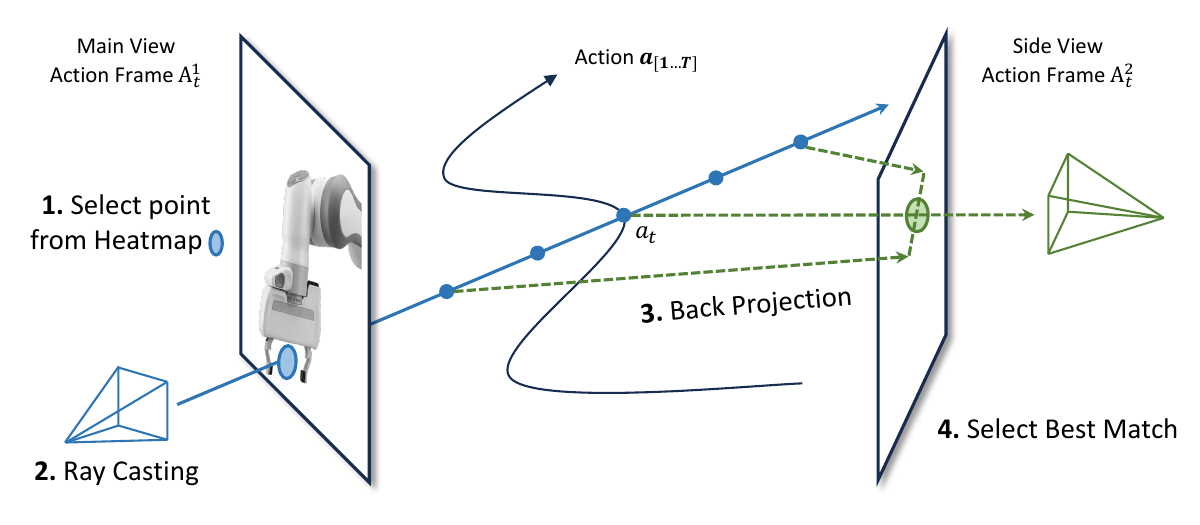}
    \caption{\textbf{Action images decoding.} A 2D heatmap point is selected in the main view, lifted to 3D by ray casting and side-view matching, and repeated for all semantic points to recover the original 7-DoF action.}
    \label{fig:action_decode}
\end{figure}

\subsection{Action Images Decoding}
\label{sec:action_decode}

A useful action representation should not only be easy to generate, but easy to decode back into continuous robot control. We therefore design a simple decoding method that maps the generated multi-view action videos back to the original 7-DoF action. The decoder first reads the gripper state directly from the blue channel, then reconstructs the underlying 3D semantic points from multi-view heatmaps, and finally converts them back into the action vector. In this way, the same unified video-space representation of observation and action can be used both for generation and for control.

\noindent\textbf{Decoding gripper openness.}
The blue channel stores both one projected semantic point and the gripper openness, where the latter is written into low-response background regions. Let $\mathbf{A}^{(v)}_t(:,:,3)$ denote the blue channel of the action image at time $t$ and view $v$. We estimate gripper openness by averaging only the low-response pixels:
\begin{equation}
\hat{g}_t
=
\frac{1}{0.25}
\cdot
\frac{1}{|\Omega_t|}
\sum_{(i,j,v)\in\Omega_t} \mathbf{A}^{(v)}_t(i,j,3),
\ \ \ \
\Omega_t = \{(i,j,v)\mid \mathbf{A}^{(v)}_t(i,j,3) < 0.25\}.
\end{equation}

\noindent\textbf{Reconstructing 3D semantic points from multi-view heatmaps.}
For the remaining action information, we decode each semantic point from its corresponding heatmap using a simple multi-view geometric procedure. As illustrated in Fig.~\ref{fig:action_decode}, we first select a 2D point from the heatmap in the main view by weighted averaging:
\begin{equation}
\hat{\mathbf{u}}^{(1)}_t
=
\frac{
\sum_{i,j}
\mathbf{H}^{(1)}_t(i,j)
\begin{bmatrix}
i+0.5, & j+0.5
\end{bmatrix}^{\!\top}
}{
\sum_{i,j} \mathbf{H}^{(1)}_t(i,j)
}.
\end{equation}
where $\mathbf{H}^{(1)}_t \in [0,1]^{H\times W}$ is the heatmap in the main view. This gives the centroid of the heat distribution and serves as the 2D anchor point for decoding.

Starting from this point, we cast a ray from the main-view camera center through $\hat{\mathbf{u}}^{(1)}_t$, and sample a set of candidate 3D points along the ray between a near plane and a far plane. Each candidate is then projected into the side view, where it is scored against the corresponding side-view heatmap. We choose the 3D point whose projection best matches the side-view response. Concretely, if $\{\mathbf{x}_{t,k}\}_{k=1}^{K}$ denotes the sampled 3D candidates along the ray, then we select
\begin{equation}
\hat{\mathbf{x}}_t
=
\arg\max_{\mathbf{x}_{t,k}}
\;
\mathbf{H}^{(2)}_t\!\left(\pi^{(2)}_t(\mathbf{x}_{t,k})\right),
\end{equation}
where $\pi^{(2)}_t(\cdot)$ is the side-view projection and $\mathbf{H}^{(2)}_t$ is the side-view heatmap. In practice, this procedure is repeated for each semantic point heatmap in the action image, yielding a set of reconstructed 3D points. The main view provides the image-space anchor for ray casting, while the side view resolves the depth ambiguity by selecting the best match along the ray.

\noindent\textbf{From reconstructed points back to 7-DoF action.}
Once the semantic 3D points are reconstructed, the original action can be recovered directly. Let $\hat{\mathbf{q}}^{\text{pos}}_t$, $\hat{\mathbf{q}}^{\text{up}}_t$, and $\hat{\mathbf{q}}^{\text{normal}}_t$ denote the decoded 3D points. We recover the position as $\hat{\mathbf{p}}_t=\hat{\mathbf{q}}^{\text{pos}}_t$, define $\hat{\mathbf{e}}^x_t=\text{norm}(\hat{\mathbf{q}}^{\text{up}}_t-\hat{\mathbf{q}}^{\text{pos}}_t)$ and $\hat{\mathbf{e}}^z_t=\text{norm}(\hat{\mathbf{q}}^{\text{pos}}_t-\hat{\mathbf{q}}^{\text{normal}}_t)$, then obtain $\hat{\mathbf{e}}^y_t=\hat{\mathbf{e}}^z_t\times\hat{\mathbf{e}}^x_t$, from which the end-effector orientation $\hat{\boldsymbol{\theta}}_t$ is determined. The final decoded action is $\hat{\mathbf{a}}_t=[\hat{\mathbf{p}}_t,\hat{\boldsymbol{\theta}}_t,\hat{g}_t]$.

\noindent\textbf{Discussion.}
When the predicted heatmaps are accurate, the remaining decoding error is dominated not by representation mismatch, but by discretization. In particular, the 3D reconstruction accuracy is mainly determined by (i) the sampling interval along the ray, which controls depth precision, and (ii) the spatial resolution of the heatmaps, which controls localization precision in image space. As a result, the information loss introduced by the action-frame parameterization is minor and predictable: finer ray sampling and higher image resolution directly improve the fidelity of the decoded action.

\subsection{Training Unified World Action Model}
\label{sec:train_uawm}

\begin{figure}[t]
    \centering
    \includegraphics[width=0.95\linewidth]{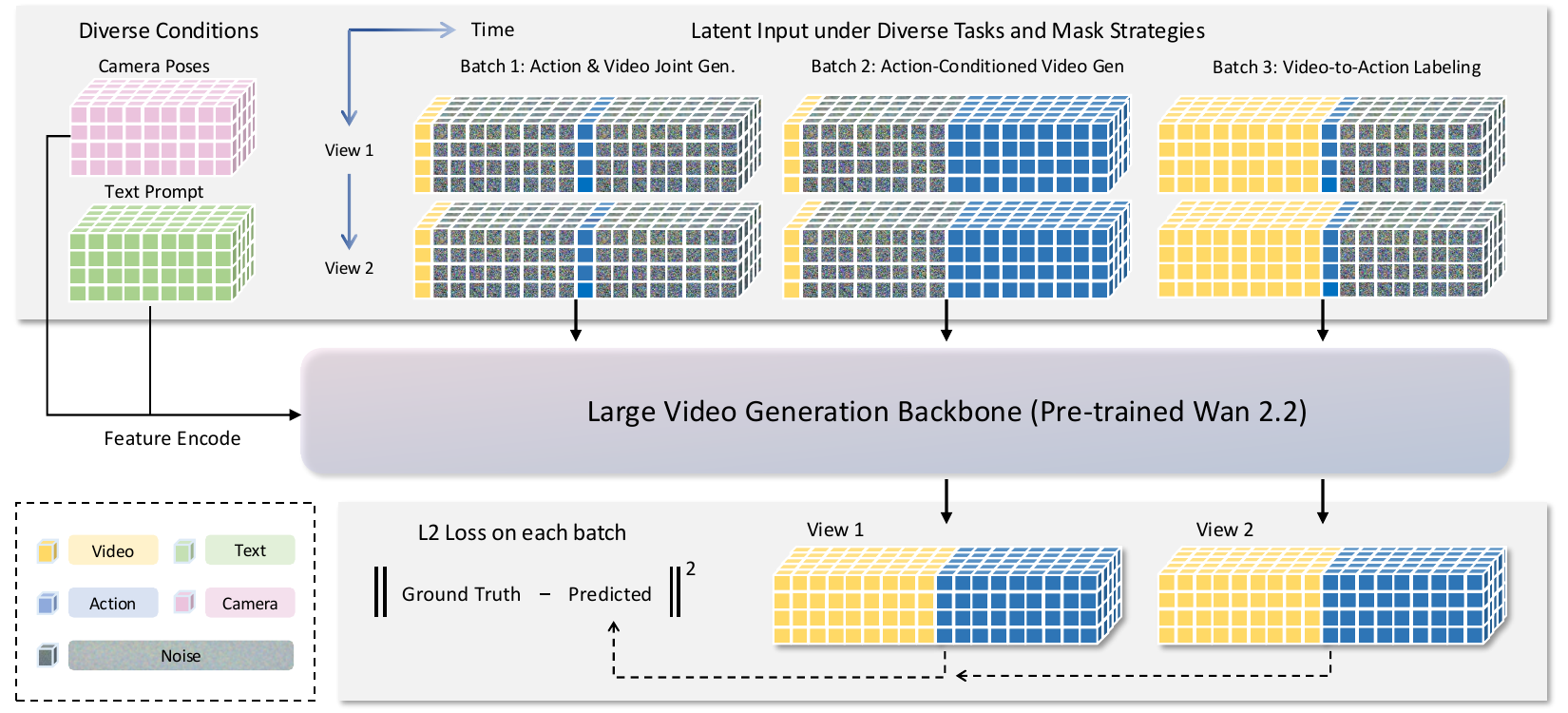}
    \caption{\textbf{Unified world-action model training.} Multi-view video and action latents are packed with text and camera conditions, and trained under diverse mask strategies.}
    \label{fig:arch}
\end{figure}

With robot actions represented as interpretable action images, control becomes a pixel-grounded visual signal rather than an abstract low-dimensional vector. This converts action modeling into the same video-space problem as observation modeling, yielding a unified video-space representation of observation and action. As shown in Fig.~\ref{fig:arch}, we build a unified world action model by fine-tuning a large pretrained video generator (Wan 2.2~\cite{wang2025wan}) to jointly model multi-view robot videos and multi-view action videos under diverse supervision patterns.

\textbf{Multi-view video-action tokenization and packing.}
For each camera view $v$, we have an RGB observation clip $\mathbf{V}_{1:T}^{(v)}\in[0,1]^{T\times H\times W\times 3}$ and the aligned action-frame clip $\mathbf{A}_{1:T}^{(v)}\in[0,1]^{T\times H\times W\times 3}$. We first encode both streams into the backbone latent space by the 3D-VAE~\cite{kingma2013auto, wang2025wan}, and then concatenate them temporally to form a single input sequence
\begin{equation}
\mathbf{X}_v \;=\; \big[\;\mathbf{V}_{1:T}^{(v)}\;,\;\mathbf{A}_{1:T}^{(v)}\;\big] \in \mathbb{R}^{(2T)\times h\times w\times c},
\end{equation}
so that the model observes, for each view, a unified timeline of \texttt{(robot video $\rightarrow$ action video)}. Multi-view data are processed with shared weights across views, enabling consistent cross-view learning while preserving per-view conditioning.

\textbf{Unified training via multiple mask strategies.}
To support multiple tasks with a single model, we adopt a multiple mask strategy in the latent token space (\Cref{fig:arch}). Concretely, we randomly sample masks over the concatenated latent sequence $\mathbf{X}_k$ to instantiate different training objectives within the same diffusion-style denoising framework:
\textbf{1)} Action \& video joint generation. We mask both $\mathcal{V}$ and $\mathcal{A}$ tokens except for the first observation frame, and ask the model to generate them jointly conditioned on text and camera inputs.
\textbf{2)} Action-conditioned video generation. We keep $\mathcal{A}$ visible while masking $\mathcal{V}$, training the model to synthesize future visual observations consistent with provided actions.
\textbf{3)} Video-to-action labeling. We keep $\mathcal{V}$ visible while masking $\mathcal{A}$, training the model to infer action images from the input video.
\textbf{4)} Video-only generation. For data without usable action, we train the model with video tokens only, using the same denoising objective to model future observations.
This masking scheme turns the same backbone into a unified world model that can switch behaviors by changing which token subsets are observed vs.\ predicted, improving generalization across settings and downstream usages.

Beyond masking-based supervision, our unified model also supports camera-controlled generation, which also helps maintain multi-view consistency. Following ReCamMaster~\cite{bai2025recammaster}, we inject camera plucker embedding~\cite{plucker1865xvii} into the backbone as
$\mathbf{F}_i \;=\; \mathbf{F}_o \;+\; E_c(\mathbf{cam}_{t}),$
where $E_c$ is a lightweight convolutional encoder, $\mathbf{F}_o$ is the output of the spatial-attention layer, and $\mathbf{F}_i$ is the input to the subsequent 3D-attention layer.

\begin{table}[t]
    \centering
    \small
    \setlength{\tabcolsep}{4pt}
    \renewcommand{\arraystretch}{1.15}
    \caption{Summary of dataset for unified world action model training.}
    \begin{tabular}{lcc|cc|cc}
        \toprule
        \textbf{Dataset} & \textbf{\#Traj.} & \textbf{\#Views} & \textbf{Real} & \textbf{Action Ann.} & \textbf{Cam. Calib.} & \textbf{Cam. Motion} \\
        \midrule
        DROID    & 80k & 2     & \cmark & \cmark & \cmark & Static \\
        RLBench  & 180k & 4     & \xmark & \cmark & \cmark & Diverse \\
        BridgeV2 & 30k & 1-4 & \cmark & \cmark & \xmark & Static \\
        \bottomrule
    \end{tabular}
    \label{tab:datasets}
\end{table}

\textbf{Optimization objective.}
We fine-tune the pretrained backbone using a flow matching~\cite{lipman2022flow} objective on the masked latent tokens. The target velocity is defined as $\mathbf{v}=\boldsymbol{\epsilon}-\mathbf{X}$. We then minimize an $L_2$ loss between the predicted and target velocities over the masked tokens:
\begin{equation}
\mathcal{L} \;=\; \mathbb{E}\Big[\big\| M \odot \big(\mathbf{v} - \mathbf{v}_\theta(\mathbf{X}, \mathcal{T}, \mathbf{cam})\big)\big\|_2^2\Big],
\end{equation}
where $M$ is the mask, $\mathcal{T}$ is the text input, and $\mathbf{cam}$ is the camera condition. This yields a single unified model that learns the coupled dynamics of visual observations and actions across multiple views.

\textbf{Training datasets.}
Training a unified world action model requires large-scale data, but this is challenging in robotics: multi-view datasets are limited, and datasets with well-aligned action and camera annotations are even rarer. We therefore train on a mixture of RLBench~\cite{james2020rlbench}, DROID~\cite{khazatsky2024droid}, and BridgeV2~\cite{walke2023bridgedata}, which provide complementary supervision as shown in Tab.~\ref{tab:datasets}. DROID offers the most complete real-robot annotations, but its camera calibration is often noisy or incomplete in practice, so we filter out low-quality samples. RLBench, although more toy-like than real-world data, provides highly accurate action and camera signals from simulation; we improve its visual diversity with Robot-Colosseum~\cite{pumacay2024colosseum} background augmentation. BridgeV2 contains high-quality real-world videos, but lacks camera labels and action-camera alignment. We estimate camera annotations with VGGT~\cite{wang2025vggt} and use BridgeV2 for video-only generation.

\section{Experiments}

\subsection{Text-Controlled Action \& Video Joint Generation}

\begin{table}[t]
\centering
\setlength{\tabcolsep}{4pt}
\caption{Zero-shot evaluation results on RLBench and Real-world settings.}
\resizebox{0.96\linewidth}{!}{
\begin{tabular}{lcccc|ccccc}
\toprule
\multirow{3}{*}{Methods} & \multicolumn{4}{c}{RLBench} & \multicolumn{5}{c}{Real}\\
\cmidrule(lr){2-10}
 & 
\makecell{\small\texttt{pick}\\\small\texttt{cup}} & 
\makecell{\small\texttt{reach}\\\small\texttt{target}} & 
\makecell{\small\texttt{close}\\\small\texttt{drawer}} & 
\makecell{\small\texttt{close}\\\small\texttt{laptop}}
&
\makecell{\small\texttt{Place}\\\small\texttt{Cup}}
&
\makecell{\small\texttt{Pick}\\\small\texttt{Unseen Toy}}
&
\makecell{\small\texttt{Pick}\\\small\texttt{Tissue}}
&
\makecell{\small\texttt{Close}\\\small\texttt{Drawer}}
&
\makecell{\small\texttt{Close}\\\small\texttt{Box}}
\\
\midrule
MV-Policy & 0 & 0 & 0 & 0 & 0 & 0 & 0 & 0 & 0 \\
${\pi_{0.5}}$ & 0 & 5 & 35 & \textbf{20} & 5 & 0 & 0 & 0 & 0 \\
MolmoAct & 20 & 5 & 10 & 0 & 10 & 5 & 5 & 5 & 0 \\
TesserAct & 0 & 0 & 0 & 0 &  0 & 0 & 0 & 0 & 0 \\
Cosmos-Policy & 0 & 5 & 20 & 0 & 0 & 0 & 0 & 0 & 0 \\
\midrule
Ours & \textbf{30} & \textbf{60} & \textbf{50} & 15 & \textbf{40} & \textbf{20} & \textbf{15} & \textbf{45} & \textbf{10} \\
\bottomrule
\end{tabular}
}
\label{tab:exp-rlbench-zeroshot}
\end{table}

We treat text-controlled action and video joint generation as the primary evaluation setting of this paper. Given a language instruction and the initial multi-view observations, the model jointly generates future robot videos and corresponding multi-view action videos, from which executable controls are obtained by decoding the predicted action images. Unless otherwise specified, all experiments in this subsection are conducted in the multi-view setting under one-trial open-loop evaluation. This is a particularly challenging setting, since the model must complete the task from a single forward prediction without online replanning, making the results directly reflect the quality and generalization ability of the learned pixel-grounded action representation.

\textbf{Zero-shot policy results.}
We compare against several representative robot policy baselines, including MV-Policy~\cite{chi2025diffusion}, $\pi_{0.5}$~\cite{intelligence2025pi_}, MolmoAct~\cite{lee2025molmoact}, TesserAct~\cite{zhen2025tesseract}, and Cosmos-Policy~\cite{kim2026cosmos}. MV-Policy is a multi-view extension of Diffusion Policy that encodes images from multiple camera views.
$\pi_{0.5}$ and MolmoAct are VLA-style baselines. For $\pi_{0.5}$, we use the base checkpoint and augment the model with an MLP that injects camera parameters into the VLM. MolmoAct is a reasoning-based model that can predict 2D trajectories on images; we leverage this capability by querying trajectories in multiple views and lifting them into 3D motion.
TesserAct and Cosmos-Policy are world-model-based baselines. For fair comparison, we reproduce both by fine-tuning the same Wan~2.2~\cite{wang2025wan} video backbone on our training set.

For evaluation, we use task success rate as the metric in both simulation and real-robot settings. The zero-shot setting differs across environments. In RLBench~\cite{james2020rlbench}, the evaluated tasks may appear in other datasets, but these specific tasks are fully removed from the RLBench training split; the robot arm and environment are seen.
In the real-world setting, the objects, environments, and robot arm (xArm) are all unseen. Across all settings, the language instructions are similar in form to those seen during training.
As shown in Tab.~\ref{tab:exp-rlbench-zeroshot}, our method delivers the best overall zero-shot performance across simulation and real-world tasks. The improvement is most evident under strong distribution shift, supporting our claim that interpretable action images and a pixel-grounded action representation lead to a more generalizable zero-shot policy.

\begin{table}[t]
\centering
\small
\caption{RLBench in-domain tasks evaluation}
\resizebox{0.98\linewidth}{!}{
\begin{tabular}{lccccccccc|c}
\toprule
Methods &
\makecell{\small\texttt{close}\\\small\texttt{box}} & 
\makecell{\small\texttt{close}\\\small\texttt{door}} & 
\makecell{\small\texttt{open}\\\small\texttt{door}} & 
\makecell{\small\texttt{phone}\\\small\texttt{base}} & 
\makecell{\small\texttt{open}\\\small\texttt{bottle}} & 
\makecell{\small\texttt{close}\\\small\texttt{drawer}} & 
\makecell{\small\texttt{open}\\\small\texttt{oven}} & 
\makecell{\small\texttt{open}\\\small\texttt{jar}} & 
\makecell{\small\texttt{wipe}\\\small\texttt{desk}} & \texttt{ Avg. }\\
\midrule
MV-Diffusion Policy & 20 & 40 & \textbf{15} & \textbf{20} & 5 & 50 & \underline{10} & 0 & 0 & 17.8\\
MomolAct (zeroshot) & 5 & 10 & 0 & 0 & 5 & 10 & 0 & 0 & 0 & 3.3\\
$\pi_{0.5}$ & 10 & 0 & 5 & 5 & \textbf{45} & 65 & 0 & 0  & 0 & 14.4 \\
TesserAct & 40 & 25 & 5 & 15 & 20 & 70 & 5 & 5 & 0 & \underline{20.6}\\
Cosmos-Policy & 40 & 15 & 0 & 15 & 30 & \textbf{80} & 0 & 0 & 0 & 20.0\\
\midrule
Ours & \underline{55} & \underline{60} & 0 & 0 & 5 & 60 & 5 & 0 & 0 & \underline{20.6} \\
w/ action head & \textbf{80} & \textbf{65} & \textbf{15} & \textbf{20} & \underline{40} & \textbf{80} & \textbf{15} & \textbf{5} & \textbf{10} & \textbf{36.7}\\
\bottomrule
\end{tabular}
}
\label{tab:exp-rlbench}
\end{table}

\textbf{RLBench in-domain results.}
We next evaluate the same model on in-domain RLBench tasks, using the same baselines and the metrics as above. Besides the reconstruction-based decoder that recovers actions from generated action images, we also consider an optional learned action head on top of the unified backbone. Specifically, we attach a lightweight MLP that takes as input the output video latents, camera parameters, and decoded actions and observations, and train it to directly regress the continuous 7-DoF action sequence.
This head is not required for our main zero-shot policy claim; rather, it is introduced to test whether the learned representation can support improved decoding.

As shown in Tab.~\ref{tab:exp-rlbench}, our method remains competitive on in-domain RLBench tasks even under the same challenging setting. Moreover, adding the optional action head brings substantial gains, especially on precision-sensitive tasks, showing that the action images can support stronger action decoding when additional supervision is available.

\textbf{Joint generation quality.}
Unlike the policy evaluations above, this experiment focuses on how accurately the model predicts both future robot videos and the corresponding actions. We compare against world models, including Cosmos-Predict~\cite{ali2025world}, Cosmos-Policy~\cite{kim2026cosmos} and TesserAct~\cite{zhen2025tesseract}. For video quality, we use PSNR and SSIM to measure pixel-level fidelity and structural similarity, with FVD and LPIPS to evaluate perceptual and temporal realism. For action quality, we report both 2D and 3D trajectory error. Since all compared models, except ours, directly predict 3D actions, we additionally project the outputs using camera parameters to obtain 2D errors. Video generation is evaluated on in-domain RLBench, Bridge, and DROID, while action metrics are evaluated on RLBench only. As shown in Tab.~\ref{fig:joint}, our method outperforms prior world-model baselines on all video metrics while maintaining action accuracy.

\begin{table}[tbp]
\centering
\setlength{\tabcolsep}{4pt}
\caption{Video-and-Action Joint Generation Quality. ($\dag$ denotes a zero-shot model.).}
\resizebox{0.98\linewidth}{!}{
\begin{tabular}{l|cccc|cc}
\toprule
\multirow{2}{*}{Models} & \multicolumn{4}{c|}{Video}    & \multicolumn{2}{c}{Action} \\
 &
  PSNR $\uparrow$&
  \multicolumn{1}{c}{SSIM (\%) $\uparrow$} &
  \multicolumn{1}{c}{FVD $\downarrow$} &
  \multicolumn{1}{c|}{LPIPS $\downarrow$} &
  2DErr $\downarrow$ &
  \multicolumn{1}{c}{3DErr $\times 10^3$ $\downarrow$} \\ \midrule
Cosmos-Predict2.5-14B$^\dag$ & 17.92 & 50.77 & 208.65 & 0.409 & - & - \\
Cosmos-Policy & 18.29 & 53.41 & 192.58 & 0.418 & 2.11 & 19.4 \\
TesserAct & 20.83 & 59.20 & 154.38 & 0.351 & 1.84 & 19.0 \\
TesserAct-RGB & 20.31 & 60.19 & 147.83 & 0.372 & \textbf{1.55} & 14.2 \\
\midrule
Ours & \textbf{23.48} & \textbf{78.62} & \textbf{143.74} & \textbf{0.209} & 1.61 & \textbf{12.2} \\
\bottomrule
\end{tabular}
}
\label{fig:joint}
\end{table}

\subsection{Additional Unified-Model Capabilities}

\begin{table}[tbp]
\centering
\begin{minipage}[t]{0.51\linewidth}
    \centering
    \caption{Action-cond. video quality.}
    \label{tab:action-cond}
    \setlength{\tabcolsep}{4pt}
    \resizebox{\linewidth}{!}{
    \begin{tabular}{l|cccc}
    \toprule
    Models & PSNR $\uparrow$ & SSIM (\%) $\uparrow$ & LVD $\downarrow$ & LPIPS (\%) $\downarrow$\\ 
    \midrule
    Tora & 19.76 & 52.43 & 187.41 & 39.62 \\
    Ours & \textbf{31.35} & \textbf{67.16} & \textbf{115.02} & \textbf{21.78} \\
    \bottomrule
    \end{tabular}
    }
\end{minipage}
\begin{minipage}[t]{0.48\linewidth}
    \centering
    \caption{Video-to-action labeling results.}
    \label{tab:action-label}
    \setlength{\tabcolsep}{4pt}
    \resizebox{\linewidth}{!}{
    \begin{tabular}{l|ccc}
    \toprule
    Models & Traj Err $\downarrow$ & Jaccard @ $\uparrow$ 4 & Avg. Jaccard $\uparrow$ \\ 
    \midrule
    TAPIR & 14.80 & 40.26 & 29.77 \\
    CoTracker & 12.91 & 46.15 & 31.20 \\
    Ours & \textbf{5.785} & \textbf{64.92} & \textbf{46.71} \\
    \bottomrule
    \end{tabular}
    }
\end{minipage}
\end{table}
\begin{figure}[t]
    \centering
    \includegraphics[width=0.98\linewidth]{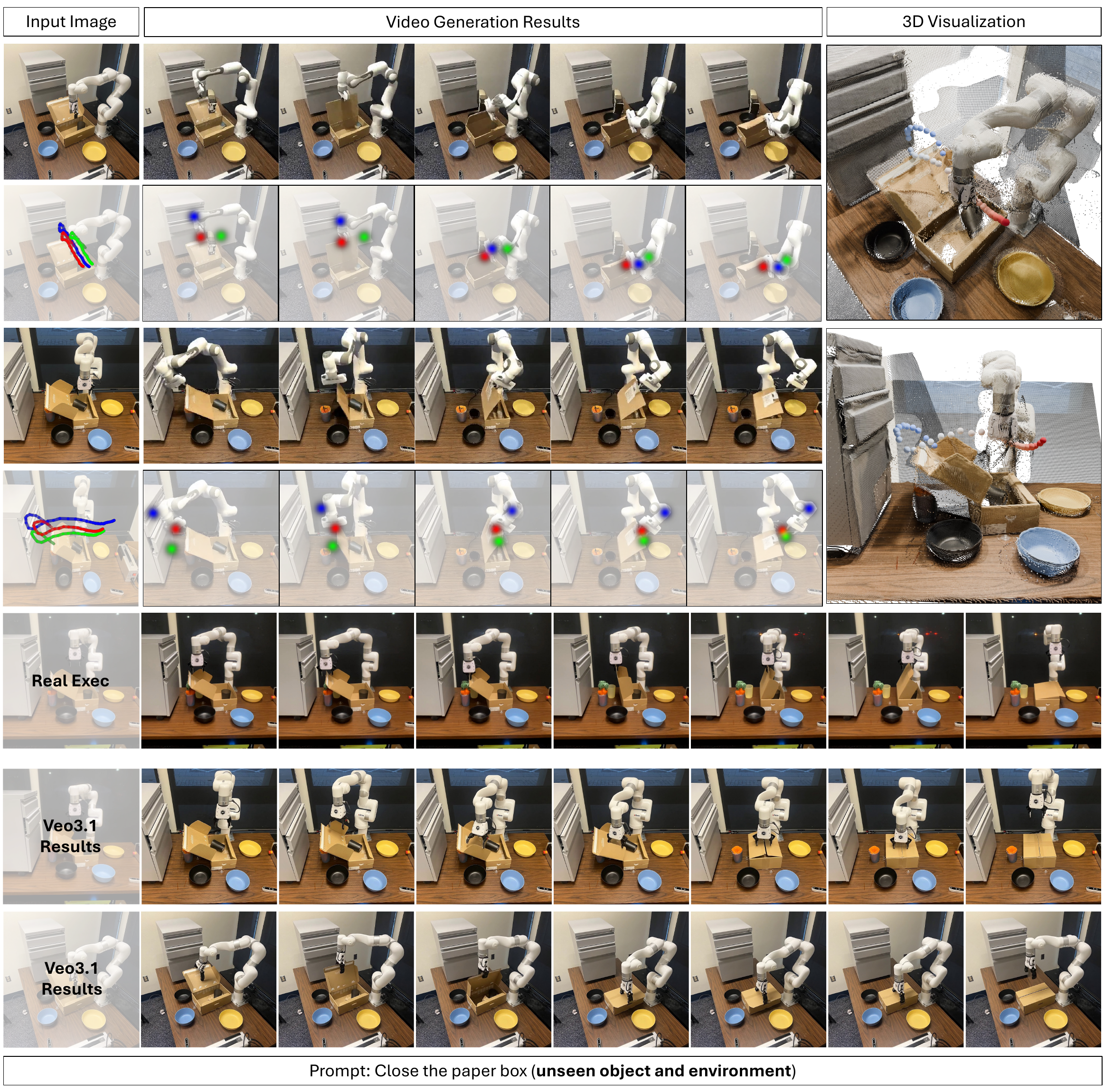}
    \caption{\textbf{Real-world zero-shot rollouts on xArm robot.} From left to right, we show the input observation, generated future video frames with predicted action-image trajectories, and the reconstructed 3D visualization. The results demonstrate that our model can generalize to unseen real-world objects and environments, while producing executable action predictions that are consistent with the generated visual outcomes.}
    \vspace{-10mm}
    \label{fig:qualitative_results_fr3m}
\end{figure}

\textbf{Action-conditioned video generation.}
This task tests whether the model can generate future robot videos when the action sequence is given.
We compare with Tora~\cite{zhang2025tora}, a 2D trajectory-conditioned video generation baseline. We evaluate generation quality using standard video metrics, including PSNR, SSIM, FVD, and LPIPS. As shown in Tab.~\ref{tab:action-cond}, our method achieves better results on all metrics, suggesting that the unified video-space representation of observation and action can use action inputs more effectively for future video prediction.

\textbf{Video-to-action labeling.}
This task tests whether the model can infer action-related motion directly from input videos.
We compare with two point-tracking baselines, TAPIR~\cite{doersch2023tapir} and CoTracker3~\cite{karaev2025cotracker3}. We use standard tracking metrics, including trajectory error, Jaccard@4, and average Jaccard. As shown in Tab.~\ref{tab:action-label}, our method outperforms both baselines by a clear margin. This result shows that the pixel-grounded action representation is not only useful for control and generation, but also provides a simple way to label action from video.

\subsection{Qualitative Results}

\begin{figure}[t]
    \centering
    \includegraphics[width=0.98\linewidth]{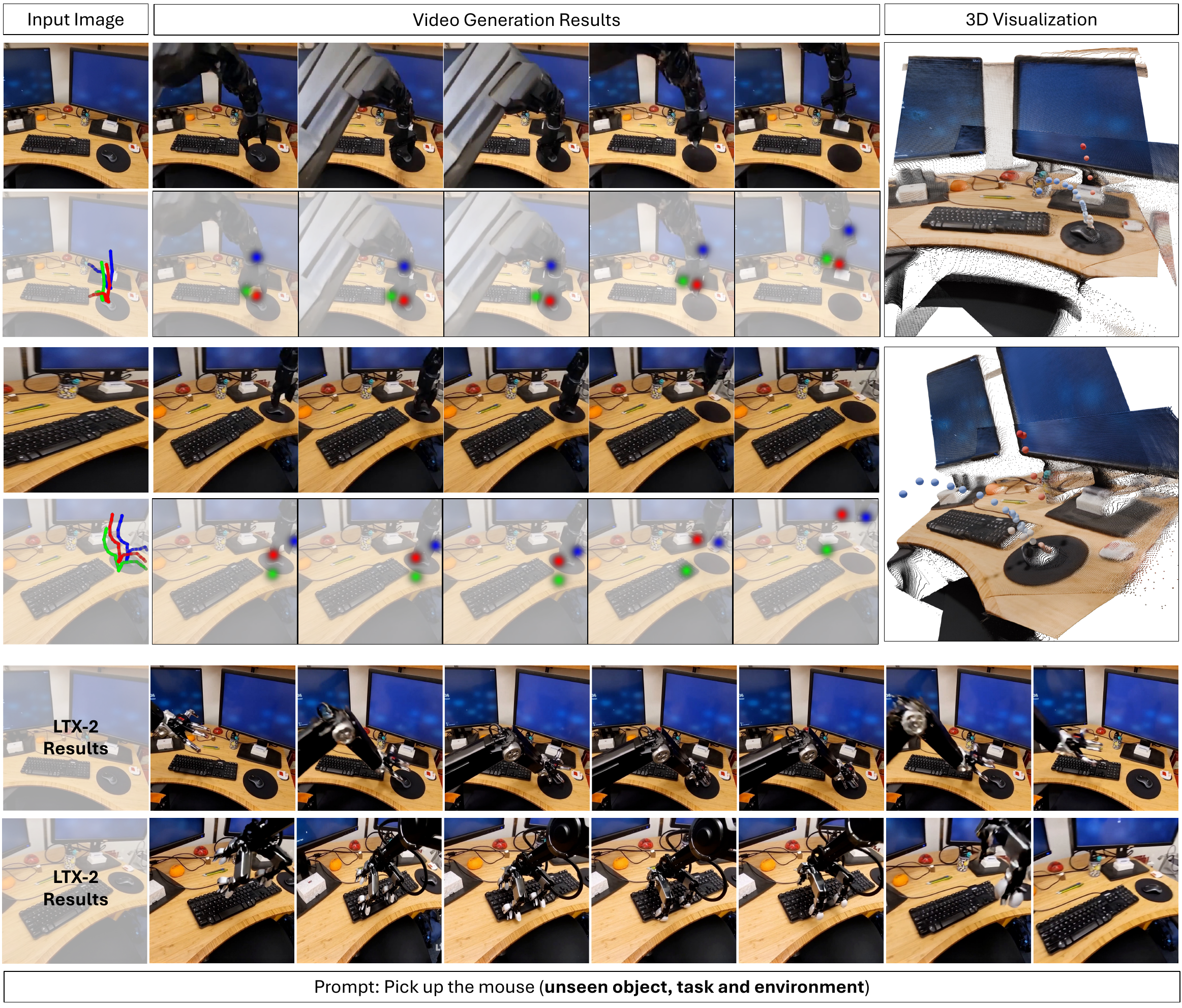}
    \caption{\textbf{Zero-shot video and action-image generation on FR3M~\cite{shen2023distilled} rooms.} This example illustrates a challenging setting with an unseen object, unseen task, and unseen environment (\texttt{pick up the mouse}). The predicted action trajectories stay aligned with the scene geometry.}
    \label{fig:qualitative_results}
\end{figure}

We first evaluate zero-shot rollouts on an xArm platform, where the objects and environment are unseen. As shown in Fig.~\ref{fig:qualitative_results}, we visualize the input image with the predicted tracking trajectories on the left. Our model first generates future observations and multi-view action images (middle), and we then decode the predicted 2D action into a 3D trajectory for point-cloud visualization (right), where the scene geometry is reconstructed by VGGT~\cite{wang2025vggt}. The 3D trajectory is colored by time, from blue (earlier) to red (later). We replay the decoded trajectory on the real robot to validate executability. Separately, we include results from a strong video-generation baseline (Veo3.1~\cite{googledeepmind_veo3_techreport_2025}) for qualitative comparison. The execution matches the generated motion, indicating that the predicted action images decode into plausible trajectories.

To further test generalization, we sample two images from the FR3M~\cite{shen2023distilled} room dataset and prompt the model to perform unseen task. Fig.~\ref{fig:qualitative_results_fr3m} compares our generations with LTX-2-Fast~\cite{lightricks_ltxstudio_2024}. Our model produces videos with more accurate localization of targets. Notably, despite lacking action supervision on BridgeV2 during training, the model still generates coherent action images, indicating that \textbf{the learned action-generation capability transfers} across datasets and domains.

\section{Conclusion}
We presented a world action model that formulates policy learning as video generation through a unified video-space representation of observation and action. Our key idea is to translate 7-DoF robot control into interpretable action images, yielding a pixel-grounded action in the form of multi-view videos. This design allows the video backbone itself to serve as a zero-shot policy model, without requiring a separate policy head or action module. The same model supports video-action joint generation, action-conditioned video generation, and action labeling under a shared generative framework. We hope this work suggests that grounding action in pixels provides a promising path toward more generalizable policy learning and robotics world modeling in a common video space.
\\[3pt]
\noindent\textbf{Limitations.}
Our current system demonstrates strong open-loop results, but has not yet been fully developed into a closed-loop policy. Fortunately, recent progress on diffusion acceleration and distillation provides a promising path to address this issue. In future work, we plan to distill our model for faster inference and integrate it into a closed-loop control pipeline.

\section{Acknowledgement}

We are extremely grateful to Zeyuan Yang, Jiaben Chen, Ziqiao Ma, Sriram Krishna, Hongxin Zhang, Zhou Xian, and Theophile Gervet for their helpful feedback and insightful
discussions.

\clearpage
\setcounter{section}{0}
\title{Action Images: End-to-End Policy Learning \\via Multiview Video Generation}

\begin{center}
\textbf{{\Large
Action Images: End-to-End Policy Learning
\\[5pt]
via Multiview Video Generation}}
\\[7pt]
{Supplementary Material}
\end{center}

\section{Implementation Details}
\textbf{Training Details.} We trained our unified world-action model by fine-tuning a pretrained Wan2.1-I2V-14B-480P~\cite{wang2025wan} backbone. The training data comprised Bridge~\cite{walke2023bridgedata}, RLBench~\cite{james2020rlbench}, and DROID~\cite{khazatsky2024droid}, sampled with mixture ratios of 0.2, 0.5, and 0.3, respectively. Each training sample contained 41 frames for a single view and a single modality; under the full two-view setting with both robot videos and action videos, this corresponds to 164 frames in total. We used a task mixture in which 85\% of samples were used for joint generation, while video-only, action-label, and action-conditioned generation each accounted for 5\%. Training was conducted on 32 A100 GPUs using DeepSpeed ZeRO~\cite{rajbhandari2020zero}, bfloat16 mixed precision, and gradient checkpointing. We used a per-device batch size of 1. The optimizer used a constant-with-warmup schedule with a learning rate of $5 \times 10^{-7}$, a warmup of 1000 steps, and gradient clipping with a maximum norm of 1.0. We trained the model for 100{,}000 optimization steps.

For camera conditioning, we followed the design of ReCamMaster~\cite{bai2025recammaster}, except that we used Pl\"ucker embeddings~\cite{plucker1865xvii} as the camera representation. The camera encoder first pooled the spatiotemporal camera features to a fixed resolution, then flattened and projected them into the model hidden dimension by a linear projector. We initialized the encoder projection to zeros and the final projector as an identity mapping, which stabilizes optimization at the beginning of training.

\textbf{Inference Details.} At inference time, we keep the input formatting and spatial-temporal configuration same with training. We use classifier-free guidance~\cite{ho2022classifier} with a scale of 10.0, and perform sampling for 50 denoising steps. In all experiments, inference is executed with 4-GPU Unified Sequence Parallelism~\cite{fang2024usp}. For in-the-wild images without camera annotations, we estimate camera extrinsics and intrinsics using VGGT~\cite{wang2025vggt}. As shown in Table~\ref{tab:inference_efficiency}, we further improve inference throughput by introducing several system-level optimizations, including CFG parallelism, VAE parallelism, caching, and \texttt{torch.compile}. With these optimizations, the video backbone reaches up to 71 FPS. We also note that although DreamZero-Flash achieves extremely fast inference, it relies on highly aggressive denoising steps, which leads to a severe degradation in video quality.

\begin{table}[ht]
\centering
\setlength{\tabcolsep}{4pt}
\caption{Inference efficiency.}
\label{tab:inference_efficiency}
\resizebox{0.98\linewidth}{!}{
\begin{tabular}{l|ccccccc}
\toprule
Models & Size & GPU & Steps & \#Frames & Image Res. & Inference Time (s) \\
\midrule
TesserAct & 5B & 1 H100 & 50 & 49 & (480, 640) & 137.5 \\
DreamZero & 14B & 1 H100 & 16 & 48 & (176, 320) & 5.7 \\
DreamZero-Flash & 14B & 2 GB200 & 1 & 48 & (176, 320) & 0.15\\
\midrule
Ours & 5B & 1 H100 & 50 & 164 & (512, 512) & 49.1 \\
+ Parallelism & 5B & 8 H100 & 50 & 164 & (512, 512) & 11.8 \\
+ Caching & 5B & 8 H100 & 16 & 164 & (512, 512) & 2.3 \\
\bottomrule
\end{tabular}
}
\end{table}

\textbf{Action Images Details.}
Following Sec.~3.1, each robot action is converted into three semantic 3D points (position, normal, and up) and projected into image space. The normal and up points are placed at a fixed distance of $0.1$ from the position point along their directions. The projected 2D points are then rasterized as Gaussian heatmaps with a standard deviation $\sigma = 0.05$ relative to the image resolution.
In practice, we observe that moderate changes to these hyperparameters do not noticeably affect performance, as long as the projected points remain within the image plane.

\section{More Zero-shot Qualitative Results}

\begin{figure}[htbp]
    \centering
    \includegraphics[width=0.98\linewidth]{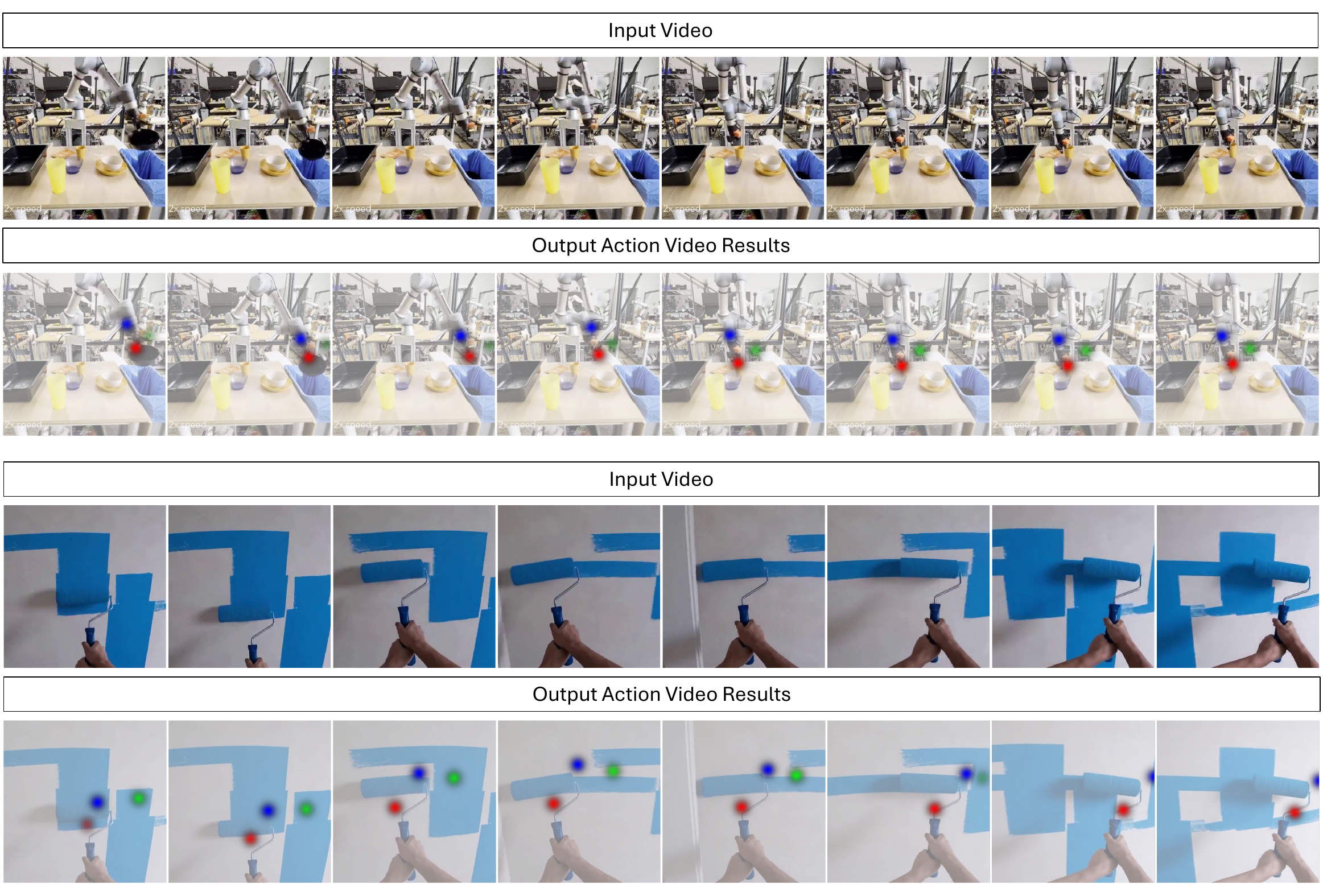}
    \caption{Action labeling results.}
    \label{fig:action-label}
\end{figure}

First, Fig.~\ref{fig:action-label} shows action labeling results given input videos, including one $\pi_0$~\cite{black2024pi_0} robot video and one Genie 3~\cite{bruce2024genie} human-hand video, demonstrating that our model can handle both.

\begin{figure}[H]
    \centering
    \includegraphics[width=0.98\linewidth]{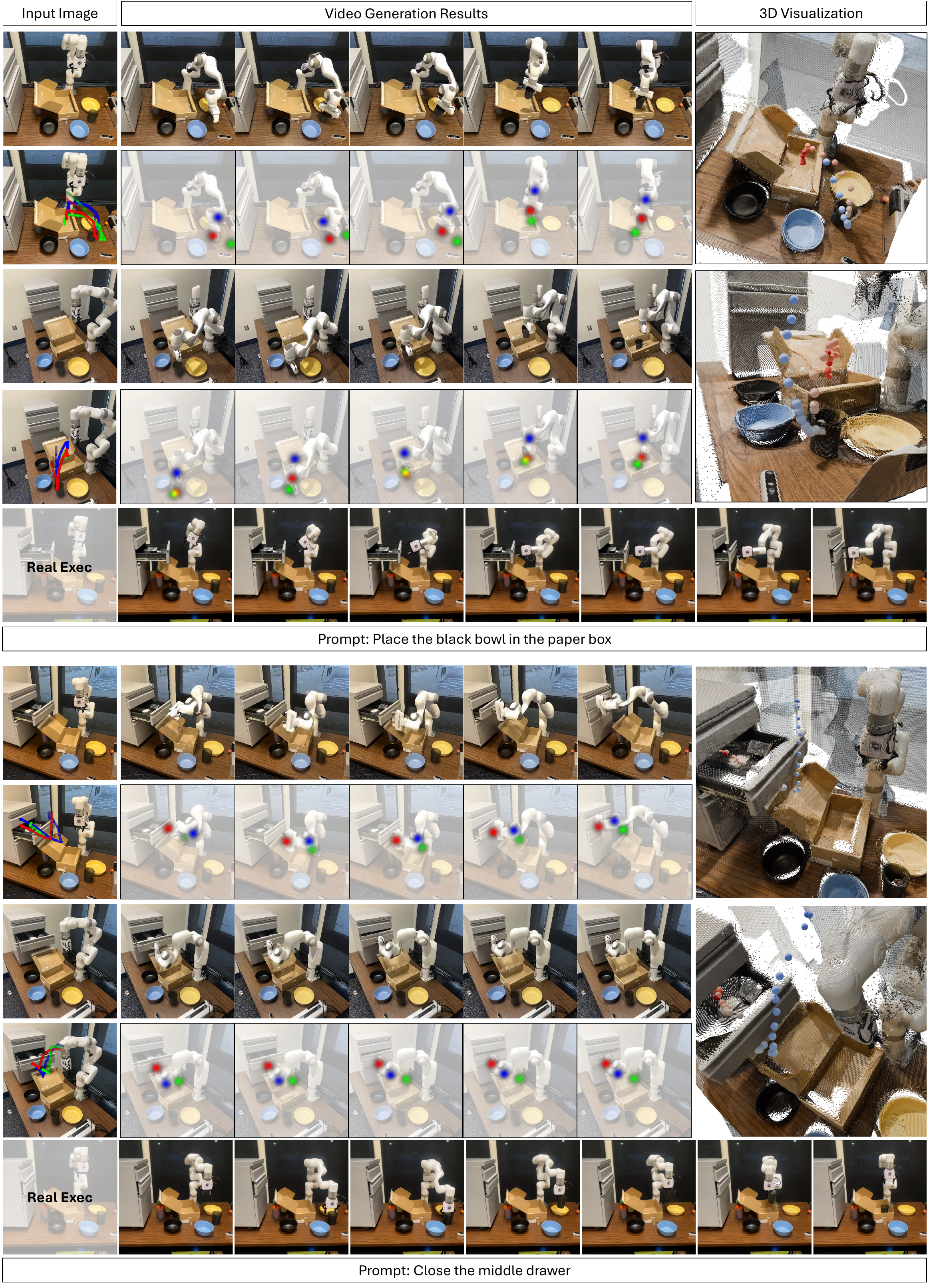}
    \caption{Additional robot manipulation results, mainly on grasping tasks.}
    \label{fig:supp-robo}
\end{figure}

\clearpage
Fig.~\ref{fig:supp-robo} provides more qualitative robot manipulation results, mainly on grasping tasks across diverse objects and scenes.

\begin{figure}[H]
    \centering
    \includegraphics[width=0.98\linewidth]{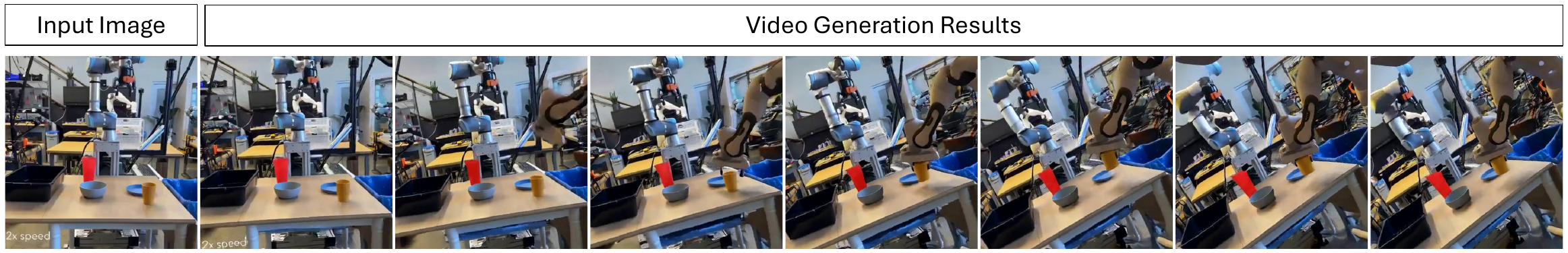}
    \caption{Camera control results in complex scenes from the $\pi_0$ website.}
    \label{fig:camera-control}
\end{figure}

We then show camera control results in Fig.~\ref{fig:camera-control}, where the model is given an input image and a task, and generates videos with controlled viewpoint changes in complex scenes from the Pi$_0$ website.

\begin{figure}[htbp]
    \centering
    \includegraphics[width=0.98\linewidth]{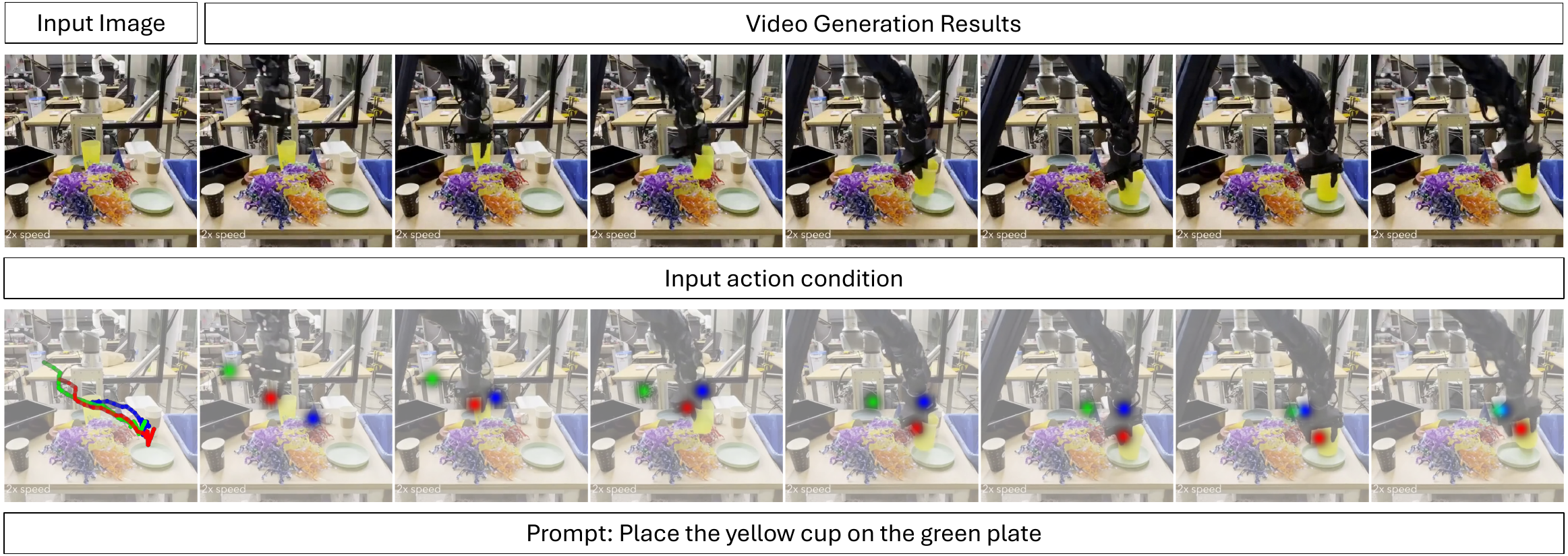}
    \caption{Action-conditioned generation results.}
    \label{fig:action-cond}
\end{figure}

Finally, we show action-conditioned generation results in Fig.~\ref{fig:action-cond}, where we use the first frame from $\pi_0$ demo videos as input to generate future videos conditioned on actions.

\clearpage
\bibliographystyle{splncs04}
\bibliography{main}

\end{document}